\documentclass{article}

\usepackage{microtype}
\usepackage{graphicx}
\usepackage{tabularx}
\usepackage{subcaption}
\usepackage{booktabs} 
\usepackage{enumitem}
\usepackage[most]{tcolorbox}
\setitemize{noitemsep,topsep=0pt,parsep=0pt,partopsep=0pt}
\usepackage{hyperref}

\usepackage[accepted]{icml2026}

\usepackage{amsmath}
\usepackage{amssymb}
\usepackage{mathtools}
\usepackage{amsthm}

\usepackage[capitalize,noabbrev]{cleveref}

\newcommand{\eg}{\textit{e.g.},~}
\newcommand{\wrt}{\textit{w.r.t.}~}
\newcommand{\ie}{\textit{i.e.},~}

\theoremstyle{plain}

\theoremstyle{definition}

\theoremstyle{remark}

\usepackage[textsize=tiny]{todonotes}

\icmltitlerunning{Procedurally Generated Tasks for improving visual grounding in MLLMs}

\begin{document}

\twocolumn[
  \icmltitle{PGT: Procedurally Generated Tasks for improving visual grounding in MLLMs}



  \icmlsetsymbol{equal}{*}

    \begin{icmlauthorlist}
      \icmlauthor{Rim Assouel}{mila,udm,fair}
      \icmlauthor{Amir Bar}{fair}
      \icmlauthor{Michal Drozdzal}{fair,equal}
      \icmlauthor{Adriana Romero-Soriano}{mila,fair,mcgill,cifar,equal}
    \end{icmlauthorlist}
    
    \icmlaffiliation{mila}{Mila - Qu\'ebec AI Institute}
    \icmlaffiliation{udm}{Universit\'e de Montr\'eal}
    \icmlaffiliation{fair}{FAIR at Meta Superintelligence Labs}
    \icmlaffiliation{mcgill}{McGill University}
    \icmlaffiliation{cifar}{Canada CIFAR AI Chair}

  \icmlcorrespondingauthor{Rim Assouel}{assouelr@mila.quebec}
  \icmlcorrespondingauthor{Adriana Romero-Soriano}{adrianars@meta.com}


  \vskip 0.3in
]



\printAffiliationsAndNotice{}  

\begin{abstract}

Despite remarkable progress in Multimodal Large Language Models (MLLMs), these models still struggle with fine-grained understanding tasks. In this work, we propose \textbf{Procedurally Generated Tasks (PGT)}, a simple data-driven framework that serves a dual purpose: inducing fine-grained visual understanding and acting as a low-cost diagnostic tool to identify the source of perception failures. By overlaying unambiguous geometric primitives on images, PGT generate additional dense supervision that disentangles visual grounding capability from semantic priors. Extensive experiments on relational, quantitative, and 3D/depth understanding benchmarks show that PGT yields remarkable gains across diverse architectures. Instruction tuning MLLMs on LLaVA-v1.5-Instruct augmented with PGT data results in improvements of up to +20\% on the What’sUp benchmark and +13.3\% on CV-Bench-2D, while maintaining general perception capabilities. Moreover, finetuning state-of-the-art MLLMs on PGT data leads to boosts of up to +5.5\% on What’sUp and +8.3\% on CV-Bench-2D. 
These findings demonstrate that PGT effectively address the bottleneck of fine-grained perception, revealing that many spatial reasoning deficits stem from inadequate supervision signals rather than inherent architectural or resolution limitations.\looseness-1
 
\end{abstract}

\section{Introduction}
\begin{figure*}[t] 
    \centering
    \includegraphics[width=\textwidth]{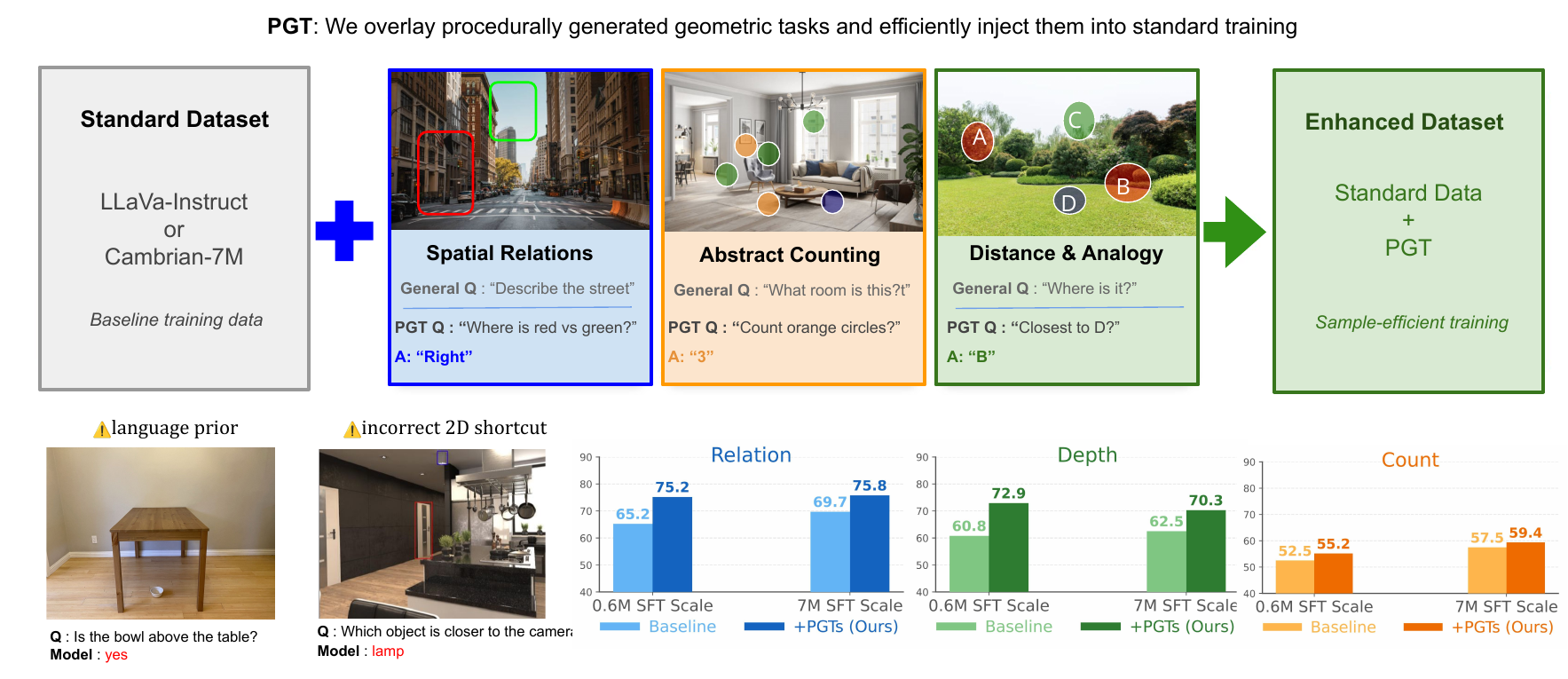}
    \caption{\textbf{Overview of PGT.} \underline{Top}: The construction of our procedurally generated data to augment instruction tuning training datasets. Abstract geometric primitives are overlaid to training data, when available. \underline{Bottom}: (Left) Examples of failure modes in fine-grained relational and spatial understanding of state-of-the-art MLLMs. In the first example the model can rely on the fact that a bowl is usually on a table and in the second example it can rely on the shortcut where object higher up as usually further from the camera (Right) PGT performance boosts in relational, quantitative, and 3D/depth understanding over the baseline when using different instruction tuning dataset sizes.\looseness-1}
    \label{fig:pgt_illustration}
\end{figure*}

Multimodal Large Language Models (MLLMs) have achieved remarkable proficiency~\citep{bai2025qwen25vltechnicalreport, zhu2025internvl3exploringadvancedtraining, li2024llavanextinterleavetacklingmultiimagevideo,tong2024cambrian1fullyopenvisioncentric,yu2025perceptionr1pioneeringperceptionpolicy, huang2025visionr1incentivizingreasoningcapability} in high-level semantic tasks, spanning image captioning, visual question answering, and open-ended dialogue. However, these strong general perception capabilities often mask a fundamental deficiency in fine-grained spatial intelligence tasks~\citep{liu2025spatial, tong2024cambrian1fullyopenvisioncentric, fu2024blinkmultimodallargelanguage} that humans can solve effortlessly. While models can eloquently describe the semantic content of a scene, they frequently falter on basic compositional reasoning tasks such as precisely counting overlapping objects, determining relative depth, or distinguishing between "left" and "right" in complex arrangements~\citep{assouel2025objectcentricbindingcontrastivelanguageimage, kamath2023whatsupvisionlanguagemodels}.\looseness-1


To mitigate these deficits, recent work has largely bifurcated into two approaches. The first involves model-centric improvements, which seek to improve perception through visual representations enhancements —such as integrating higher-resolution vision encoders~\citep{li2024llavanextinterleavetacklingmultiimagevideo}, employing a mixture of Vision Transformers (ViTs)~\citep{tong2024cambrian1fullyopenvisioncentric, karamcheti2024prismaticvlmsinvestigatingdesign} to capture multi-scale visual features, or distilling visual information from multiple experts encoders~\citep{yoon2025visualrepresentationalignmentmultimodal}. The second approach relies on data-centric scaling, either utilizing massive and costly human-annotated datasets or implementing complex multi-stage training protocols~\citep{li2025spatialladderprogressivetrainingspatial,sarch2025groundedreinforcementlearningvisual, huang2025visionr1incentivizingreasoningcapability, meng2025mmeurekaexploringfrontiersmultimodal, yu2025perceptionr1pioneeringperceptionpolicy}. Common strategies in this domain include generating \emph{thinking traces} interleaved with coordinates grounding~\citep{sarch2025groundedreinforcementlearningvisual} or enforcing auxiliary objectives like object enumeration. While effective, these methods incur high annotation bottlenecks and computational overhead, often overfitting to specific domain~\citep{liu2025spatial} distributions without fully solving the underlying grounding problem.\looseness-1

Meanwhile, recent work \citep{fu2024blinkmultimodallargelanguage} demonstrates that the failure in easy vision-centric tasks does not stem from a lack of visual representation in the encoder, but rather from insufficient training incentives to override linguistic priors~\citep{yamada2024lemonspurpleconceptassociation}. Current MLLMs often bypass genuine visual grounding, behaving akin to "bag-of-words" classifiers that associate objects (\eg "sky") with spatial concepts (\eg "up") based on text statistics rather than grounded geometry~\citep{sarch2025groundedreinforcementlearningvisual, chen2026mllmsattendrelyon}. This dependence is exacerbated by the ambiguity of natural image datasets~\citep{jian-etal-2025-teaching, yang2025thinkingspacemultimodallarge}, where spatial relationships are often implicit, providing a noisy signal that discourages the model from attending to the visual latent space.\looseness-1

A theoretically efficient way to force a model to ignore real-world induced correlations (\eg "sky is blue" or "ceiling is up") is to train on fully synthetic, abstract visual reasoning tasks where such priors do not exist. However, given the immense capacity of SOTA models, it remains unclear if this efficiency comes at the cost of relevance. We ask a fundamental question: 
\textbf{Can fine-grained spatial skills efficiently transfer from abstract, non-ambiguous objects to real-world scenarios?}. To answer this, we introduce \textbf{Procedurally Generated Tasks (PGT)}, a set of synthetic tasks that use unambiguous geometric primitives—such as colored bounding boxes, labeled points, and counting markers—overlaid directly onto existing training images (see Figure~\ref{fig:pgt_illustration}). Unlike methods relying on costly human annotation, PGT automatically generates data with dense, verifiable \emph{low-cost} supervision. By forcing the model to reason about abstract primitives where language priors do not apply (\eg a green box has no semantic correlation with being "above" or "under" a red box), we compel the visual attention mechanism to engage with the actual image geometry, testing the hypothesis that these abstract skills unlock better grounding in complex semantic environments.\looseness-1

By training MLLMs on PGT data, we demonstrate a surprising phenomenon: skills learned from PGT's abstract geometric overlays transfer effectively to real-world semantic objects. For example, we show that a model trained to estimate the distance between two abstract points can better estimate the relative depth of multiple objects in a natural image. Through extensive experimentation, we demonstrate that PGT's simple, low-cost intervention yields state-of-the-art improvements across 11 diverse benchmarks. Augmenting a Llama-3-8B-based~\citep{grattafiori2024llama3herdmodels} MLLM with PGTs yields a +19.6\% gain on the What’s Up~\citep{kamath2023whatsupvisionlanguagemodels} spatial benchmark and +13.3\% on CV-Bench 2D~\citep{tong2024cambrian1fullyopenvisioncentric}. Most notably, we observe emergent 3D capabilities, with improvements of up to +9.7\% on CV-Bench 3D, despite the training signal being purely 2D. This indicates the possibility of a shared distance estimation circuit between the 2D and 3D settings. Our contributions are summarized as follows:

\begin{itemize}
\item \textbf{A Suite of Procedurally Generated Tasks:} We propose PGT, a set of abstract geometric tasks designed to disentangle visual signals from linguistic priors. (Section~\ref{sec:PGT})
\item \textbf{Controlled Multimodal Training across Diverse Backbones:} We demonstrate the generalizability of our method through a rigorous study across four distinct LLM backbones. The consistent gains suggest that PGT addresses a fundamental modality alignment bottleneck common to a wide range of architectures. (Section~\ref{sec:mm_train})\looseness-1
\item \textbf{Impact on SOTA \& Diagnostic Value:} We push the state-of-the-art by fine-tuning advanced models (including Qwen-2.5-VL and InternVL3), showing that PGT yields significant improvements even on top of strong pre-training. Furthermore, we position PGT as an complementary diagnostic signal: its ability to improve performance without architectural changes suggests it can serve as a baseline to justify complex model-centric improvements, or be combined with them for additive gains. (Section~\ref{sec :finetuning})\looseness-1\end{itemize}
\section{Related Work}
\paragraph{Causes of Finegrained Understanding Limitations in MLLMs.}
Recent work~\citep{liu2025spatial} attributes finegrained spatial understanding limitations of MLLMs to three primary root causes spanning architecture, training pipelines, and data. \citet{covert2025localityalignmentimprovesvisionlanguage,chen2025visiontransformersselfdistilledregisters,assouel2025objectcentricbindingcontrastivelanguageimage, tschannen2025siglip2multilingualvisionlanguage, ranzinger2025featsharpvisionmodelfeatures} underline the impact of leveraging vision backbones (\eg CLIP~\cite{radford2021learningtransferablevisualmodels}) optimized for global semantic alignment on the ability of MLLMs to preserve the spatial structure needed for downstream reasoning. 
~\citet{fu2025hiddenplainsightvlms, dorkenwald2024pinpositionalinsertunlocks} argue for fine-grained supervision and explicit spatial grounding during training. 
Yet, there is a lack of large-scale, high-quality datasets annotated with explicit spatial relations~\citep{kamath2023whatsupvisionlanguagemodels,liu2025spatial}. Unlike semantic pre-training data, spatially-grounded data are expensive to annotate and often restricted to synthetic domains or limited relations. Determining the root cause of the bottleneck for a specific task typically requires expensive ablation studies or architectural modifications. PGT offers a cost-efficient alternative to verify whether heavy modifications (\eg replacing the encoder or retraining with RL) are actually justified, or if the model simply lacks a clear training signal to unlock its spatial understanding capabilities.

\paragraph{Improving Finegrained Understanding in MLLMs.} Efforts have been devoted to enhancing vision backbones architectures and training objectives, either through self-supervised post-training objectives~\citep{covert2025localityalignmentimprovesvisionlanguage} that encourage the recovery of local semantic details or through the aggregation of features from multiple expert encoders (\eg DINOv2, CLIP) via specialized connectors or distillation objectives, explicitly fusing spatially rich representations with semantic embeddings to support fine-grained tasks.
Efforts have also been devoted to devising training-free approaches such as ViperGPT~\citep{surís2023vipergptvisualinferencepython} that enhance both interpretability and performance of the models by decomposing complex visual tasks into structured sequences of subgoals. By contrast, hybrid prompting strategies integrate explicit spatial markers into the reasoning stream. For example, \citet{lei2024scaffoldingcoordinatespromotevisionlanguage,izadi2025visualstructureshelpsvisual} propose to overlay coordinate matrices or visual cues directly onto images while embedding corresponding references in the text, thereby bridging the modality gap. Similarly, \citet{li2025seegroundgroundzeroshotopenvocabulary, wu2024mindseyellmsvisualizationofthought} elicit spatial reasoning of MLLMs  by visualizing their reasoning traces. Another line of work~\citep{zhou2024imageofthoughtpromptingvisualreasoning,hu2024visualsketchpadsketchingvisual} extends the Visualization-of-Thought prompting idea to be used during inference, enabling the model to further process the visual input to support its reasoning.
Finally, the focus has most recently shifted to inducing robust spatial reasoning behaviors—such as backtracking, reflection, and sequential processing—via reinforcement learning (RL). Following this line, ViGoRL~\citep{sarch2025groundedreinforcementlearningvisual} employs a visually grounded RL framework with Monte Carlo Tree Search to generate reasoning traces which anchor every step to specific image coordinates, effectively forcing the model to "point" to evidence and backtrack when necessary. SpatialLadder~\citep{li2025spatialladderprogressivetrainingspatial} adopts a progressive training curriculum -- using RL -- to transition models from basic perception to complex multi-step reasoning. Parallel to this, Visual Jigsaw~\citep{wu2025visualjigsawposttrainingimproves} leverages RL from verifiable rewards (RLVR) on temporal and spatial ordering tasks, encouraging the model to develop sequential processing capabilities and structural awareness without reliance on expensive human annotations.\looseness-1

\section{Procedurally Generated Tasks}\label{sec:PGT}

To address the limitations of MLLMs in fine-grained vision-centric reasoning, we introduce procedurally generated tasks (PGT), a set of tasks that utilize procedurally generated geometric overlays to provide dense, verifiable supervision signals to improve MLLMs. Unlike traditional data augmentation, our PGT uses these overlays as pretext tasks that require the model to perform precise spatial and quantitative reasoning without augmenting the total number of samples.\looseness-1

\subsection{Taxonomy of PGT}
We design three core tasks to target specific reasoning deficits observed in current MLLMs.\looseness-1
\vspace{-1em}

\paragraph{Spatial Relationship Understanding.}
We overlay two distinct geometric primitives—typically a green and a red bounding box—onto a real-world image. We prompt the model to solve two sub-tasks: (1) \textit{relative positioning} -- determining the spatial relationship of one box relative to the other (\eg "above", "below", "left", "right"); and (2) \textit{coordinate regression} -- providing the normalized coordinates of a specific box to enforce precise localization. See Figure~\ref{fig:pgt_illustration} (top) for an example.\looseness-1
\vspace{-1em}

\paragraph{Counting.} To decouple counting ability from semantic object recognition, we project a variable number of semi-transparent colored circles (\eg orange, purple, green, etc) onto the scene. We prompt the model to identify and count only the circles of the requested color, ignoring both the background objects and circles of other colors. See Figure~\ref{fig:pgt_illustration} (top) for an example.\looseness-1
\vspace{-1em}

\paragraph{Analogy and Relative Distance estimation.} We utilize multi-point overlays labeled with identifiers (\eg A, B, C, D) and prompt the model to solve two sub-tasks : (1) \textit{distance estimation} -- identifying which labeled point is closest to a target point (\eg "Which one is closest to D: A, B, or C?"); and (2) \textit{simple analogy} -- identifying circles that share the same properties, (\eg color) across different locations in the image. See Figure~\ref{fig:pgt_illustration} (top) for an example.\looseness-1

\subsection{Procedural Generation and Overlaying}
A primary advantage of our PGT is that the entire generation pipeline is automated and requires no human annotation. Given an image $I$ from an existing training dataset, and its associated set of question-answer pairs $\mathcal{S} = \{(q_i, a_i)\}_{i=1}^n$, we augment each data sample $X = (I, \mathcal{S}$) in the training set through the following procedure:

\begin{enumerate}
    \item \textbf{Task Selection:} We uniformly sample a task from the previously defined PGT taxonomy (\eg spatial relationship understanding, counting, or analogy and relative distance estimation). Unless stated otherwise, we sample a task for each data point. We further ablate this design choice in section~\ref{sec_abla_prop}.
    \item \textbf{Rendering:} We sample the necessary geometric primitives $\mathcal{P}$ (\eg bounding boxes, colored circles, or labeled points) with randomized parameters for location, color, and scale. We render these primitives in an overlay layer $L$ and produce a modified image $I_{PGT} = I \oplus L$, where $\oplus$ denotes overlay.
    \item \textbf{Procedural QA Generation:} We procedurally generate a new question-answer pair $(Q_{PGT}, A_{PGT})$ based on the ground-truth geometric properties of $\mathcal{P}$.
\item \textbf{Sample Integration:} We append the newly generated question-answer pair to the original set of pairs $\mathcal{S}$, resulting in an augmented question-answer pairs set $\mathcal{S}_{aug} = \mathcal{S} \cup \{(Q_{PGT}, A_{PGT})\}$, and a PGT-augmented data sample $X_{aug} = (I_{PGT}, S_{aug})$.
\end{enumerate}

This approach preserves the original semantic integrity of the image while introducing a layer of abstract reasoning at a negligible computational overhead. It is worth noting, that when having access to real data, leveraging the overlay mechanism to alter the real data avoids augmenting the dataset size—which would necessitate more training iterations. Therefore, when real data is available, we apply these tasks as a targeted augmentation to existing datasets, such as LLaVA-Instruct-1.5~\citep{liu2023visualinstructiontuning}. By projecting abstract primitives directly onto real-world scenes, we compel the model's attention mechanism to disentangle the "base" semantic layer from the "abstract" geometric layer. This prevents the model from over-relying on linguistic priors (\eg assuming a "sky" token implies an "above" relationship) and forces it to ground its reasoning in specific visual tokens because there is no shortcut to rely on. Furthermore, because the tasks are procedurally generated, we ensure that the objects of interest and their spatial configurations remain non-ambiguous and verifiable. We give further details and exact template in Appendix~\ref{app:dataset}.\looseness-1

\section{Results}
In this section, we evaluate the performance of PGT in different settings. First, we assess the impact of PGT when building MLLMs and using PGT in their instruction tuning phase. This is done in a controlled setting in subsection ~\ref{sec:mm_train}. Then, we validate the potential of PGT when finetuning state-of-the-art MLLMs in subsection~\ref{sec :finetuning}. To align our evaluation with the core focus of this work, our primary analysis targets metrics that isolate fine-grained visual understanding and spatial grounding. Specifically, we evaluate our models using a diverse range of benchmarks spanning relational understanding  
(\textit{What's up-COCO, CV-Bench-2D, VSR}), vision-centric and 3D understanding (\textit{CV-Bench 3D, TallyQA, MMVP, RealworldQA}, and single image perception subsets of \textit{BLINK}, namely \textit{spatial, depth, relational, localization}.
Finally, we include general image understanding benchmarks (\textit{MMSTAR, GQA, POPE, SEED, AI2D}) as a sanity check to ensure that our synthetic intervention effectively overrides linguistic priors without degrading the models' broader semantic understanding capabilities.

\begin{table*}[t]
\caption{Quantitative comparison of the performance achieved by instruction tuned MLLMs across diverse benchmarks. MLLMs are built leveraging a pre-trained vision backbone and varying pre-trained base LLM backbones. For each model family, we compare the baseline instruction tuning protocol (prismatic training) against the PGT-augmented (\textbf{+ PGT}) variant. Results demonstrate that PGT yields remarkable improvements in spatial, fine-grained, and 3D reasoning without degrading general perception. All values are accuracies (\%).\looseness-1}
\label{tab:main_results_granular}
\vskip 0.15in
\begin{center}
\begin{small}
\begin{sc}
\resizebox{\textwidth}{!}{\begin{tabular}{l|ccc|ccccc|ccccc}
\toprule
& \multicolumn{3}{c|}{\textbf{Relational Reasoning}} & \multicolumn{5}{c|}{\textbf{Vision-Centric / 3D}} & \multicolumn{4}{c}{\textbf{General Perception}} \\
\textbf{Base LLM} & W-Up & CV-2D & VSR & CV-3D & Tally & MMVP & RWQA& BLINK & MMSTAR & GQA & POPE &SEED & AI2D\\
\midrule
Vicuna-1.5-7B & 75.9 & 55.8 & 54.7 & 58.4 & 63.7 & 26.0 & 53.5 &  49.2 &32.7 & 64.7 & 88.0& 62.1 &52.4\\
\textbf{+ PGT} & \textbf{95.9} & \textbf{68.1} & \textbf{68.8} & \textbf{67.1} & \textbf{66.0} & \textbf{28.0} & \textbf{54.9} & \textbf{57.3}&\textbf{35.3} & 64.1 & 87.6& \textbf{64.3} &\textbf{54.5}\\
\textit{Improvement} & +20.0 & +12.3 & +14.1 & +8.7 & +2.3 & +2.0 & +1.4 & +8.1 &+2.6 & -0.6 & -0.4& +1.2&+2.1\\
\midrule
Llama-3-8B & 77.8 & 58.5 & 63.8 & 61.1 & 63.4 & 31.3 & 59.0 &58.0 & 38.7 & 67.6 & 82.3 & 65.7&57.3\\
\textbf{+ PGT} & \textbf{97.4} & \textbf{71.8} & \textbf{72.3} & \textbf{70.8} & \textbf{66.4} & \textbf{33.3} & 58.9 &\textbf{62.8} & \textbf{39.8} & 67.1 & \textbf{83.7}& \textbf{66.3}&57.5\\
\textit{Improvement} & +19.6 & +13.3 & +8.5 & +9.7 &+3.0& +1.2 & -0.1 &+4.8 & +1.1 & -0.5 & +1.4 & +0.6&+0.2\\
\midrule
Qwen-2.5-7B & 91.8 & 60.8 & 72.9 & 65.3 & 67.0 & 38.0 & 54.9 &58.9 & 40.7 & 60.0 & 79.7 & 66.2&62.0\\
\textbf{+ PGT} & \textbf{96.3} & \textbf{72.7} & \textbf{73.2} & \textbf{73.6} & \textbf{70.9} & \textbf{40.7} & \textbf{55.8} &\textbf{62.2} & \textbf{41.5} & 59.2 & 79.4 & \textbf{67.9}&\textbf{63.7}\\
\textit{Improvement} & +4.5 & +11.9 & +0.3 & +8.3 & +3.9 & +2.7 & +0.9 &+3.3 & +0.8& -0.8 & -0.3& +1.7 &+1.7\\
\midrule
Qwen-2.5-14B & 89.1 & 64.5 & 68.9 & 63.1 & 70.6 & 34.0 & 58.2 &61.6 & 41.3 & 59.1 & 86.1& 65.9&64.7 \\
\textbf{+ PGT} & \textbf{98.3} & \textbf{74.6} & \textbf{75.2} & \textbf{73.1} & 69.5 & \textbf{43.3} & \textbf{58.2} &\textbf{63.8} & \textbf{43.1} & \textbf{62.0} & \textbf{88.0} & \textbf{69.1}&66.6\\
\textit{Improvement} & +9.2 & +10.1 & +6.3 & +10.0&-0.9 & +9.3 & +0.0 &+2.2 &+1.8& +2.9 & +1.9& +3.2&+1.9 \\
\bottomrule
\end{tabular}}
\end{sc}
\end{small}
\end{center}
\vskip -0.1in
\end{table*}
\subsection{PGT for instruction tuning}\label{sec:mm_train}
We investigate the utility of our PGT data for instruction tuning of MLLMs, \ie we assume access to a pre-trained vision backbone as well as a pre-trained large language model (LLM) backbone and build a MLLM by adding an adapter between the 2 backbones. We follow the 2 training stages of~\citet{karamcheti2024prismaticvlmsinvestigatingdesign,tong2024cambrian1fullyopenvisioncentric,li2024llavanextinterleavetacklingmultiimagevideo}: we first pre-train the adapter for captioning alignment and then intruction-tune both the adapter and the LLM backbone, while keeping the vision backbone frozen. We use PGT data for instruction tuning.

\paragraph{Experimental Setup.}

We use CLIP-ViT-L/14@336px~\cite{radford2021learningtransferablevisualmodels} as vision backbone, and consider the following LLM backbones: Vicuna-1.5-7b-instruct~\cite{liu2023visualinstructiontuning}, Llama3-8b-Instruct~\cite{grattafiori2024llama3herdmodels}, Qwen2.5-7b-Instruct~\cite{qwen2025qwen25technicalreport} and Qwen2.5-14b-Instruct~\cite{qwen2025qwen25technicalreport}. We utilize the LLaVA-v1.5-Instruct dataset~\cite{liu2023visualinstructiontuning} as our primary semantic training data source and augment it with PGT for instruction tuning. All MLLM are trained for 1 epoch for captioning alignment (adapter only) followed by 1 or 2 epochs for instruction tuning (adapter and LLM). Similar to~\citet{liu2023visualinstructiontuning, karamcheti2024prismaticvlmsinvestigatingdesign}, we use a cosine learning rate schedule with a 0.1 warmup period, a global batch size of 128 and a learning rate of $2e^-5$. To evaluate the impact of our proposed PGT, we compare models trained on the original dataset against those augmented with our tasks. As described in section~\ref{sec:PGT}, the augmented dataset maintains the exact same number of training samples as the original dataset, with PGT data injected directly into the existing visual-instruction pairs. Exact hyperparameters for this section are given in Appendix~\ref{app:mllm_training_hyper}.\looseness-1

\paragraph{Results.}
Table~\ref{tab:main_results_granular} presents the results of leveraging PGT data across diverse LLM backbones. As shown in the table, PGT yields substantial performance gains in relational reasoning benchmarks as well as vision-centric benchmarks that require precise spatial, relational and quantitative reasoning. The most notable improvements are observed in the What's Up benchmark, where VLMs based on the Vicuna-1.5-7B and Llama-3-8B models demonstrate absolute increases of +20.0\% and +19.6\% respectively. This improvement of spatial relationship understanding extends to CV-Bench-2D, which witnesses a consistent boost ranging from +10.1\% to +13.3\% across all tested LLM backbones, suggesting that the localization skills learned from abstract geometric primitives transfer effectively to real-world object relationships.\looseness-1

Furthermore, although the PGT tasks are formulated in 2D, we observe an emergent improvement in 3D and fine-grained vision-centric perception; with absolute gains ranging from +8.3\% to +10.0\% on CV-Bench-3D and from +2.2\% to +8.1\%  on BLINK. We further confirm and ablate that general 2D relative distances comparisons directly help with relative depth estimation of real world objects. We further identified heuristic failures in baseline models (see Appendix~\ref{app:response_analysis}) and hypothesize that PGT are effective by reinforcing a possible shared distance estimation circuit between the 2D and 3D setting.
Finally while specialized benchmarks like VSR show sharp increases for the Vicuna baseline (+14.1\%), the framework also proves beneficial for quantitative reasoning, as evidenced by a +3.9\% improvement on TallyQA for the Qwen-2.5-7B backbone. Critically, these enhancements do not compromise the models' foundational capabilities; performance on general perception benchmarks such as MMSTAR remains stable or improves, while scores on general tasks like GQA and POPE fluctuate minimally, confirming that PGT acts as a complementary, effective training signal that improves fine-grained image understanding  without harming broader semantic and general understanding.

\subsection{Finetuning state-of-the-art MLLMs with PGT}\label{sec :finetuning}

\begin{table*}[t]
\caption{Quantitative comparison of the impact of finetuning state-of-the-art MLLMs with PGT-augmented data over baselines and specialized methods. Results demonstrate that PGT effectively improves the baselines' performance while being competitive or improving specialized methods. 
Accuracies are reported in (\%).$^*$ means that the result is reported from original paper and not recomputed with our codebase.\looseness-1}
\label{tab:sota_results}
\vskip 0.15in
\begin{center}
\begin{small}
\begin{sc}
\resizebox{\textwidth}{!}{\begin{tabular}{l|ccc|ccccc|ccccc}
\toprule
& \multicolumn{3}{c|}{\textbf{Relational Reasoning}} & \multicolumn{5}{c|}{\textbf{Vision-Centric / 3D}} & \multicolumn{4}{c}{\textbf{General Perception}} \\
\textbf{MLLM} & W-Up & VSR & CV-2D& CV-3D & Tally & RWQA & BLINK& MMVP & SEED & MMSTAR& GQA &AI2D & POPE\\
\midrule
\multicolumn{12}{l}{\textit{Improving SOTA Backbones }} \\
\midrule
LLaVA-Next-7B & 85.2&64.7 & 59.5 & 51.6 & 45.2  & 58.4 & 55.4 & 32.7 & 63.4& 34.8&\textbf66.7&64.5&\textbf{87.8}\\
\textbf{+ PGT} &  \textbf{90.7} & \textbf{70.9} & \textbf{67.8} & \textbf{59.9} & \textbf{60.1} &\textbf{60.1} & \textbf{58.8} & \textbf{38.7} & \textbf{64.3}& \textbf{37.1}& 66.5& 64.5 &86.1\\
\midrule
LLaVa-Next-Llama3-8B & 93.8 &71.8 & 62.9 & 68.3 & 59.2 &  59.5 & 63.1 & 38.7 & \textbf{68.1}&41.5&69.3&\textbf{71.5}&\textbf{87.6}\\
\textbf{+ PGT} & 93.8 &71.8 & \textbf{65.7} & \textbf{77.3} & \textbf{63.9} &  59.5 & \textbf{64.2} & \textbf{42.0} & 67.2&\textbf{45.3}& 68.1& 71.3 &86.6\\
\midrule
InternVL3-8B & 97.2 &85.2 & 81.4 & 85.7 & 77.2 &  65.2 & \textbf{74.6}& 60.7 &76.4&68.1 &66.8& 83.8 & \textbf{91.0}\\
\textbf{+ PGT} &\textbf{97.9} &  \textbf{85.3} & \textbf{82.5} & \textbf{86.0} & \textbf{77.4} & \textbf{68.5} & 74.5 & \textbf{62.7} & \textbf{77.0}&\textbf{68.5}& \textbf{68.3}& \textbf{84.0} &90.8\\
\midrule
\multicolumn{12}{l}{\textit{PGT vs. Specialized Data \& Models }} \\
\midrule
Qwen2.5-VL-3B & 93.8 & 80.4 & 70.9 & 71.5 & 66.4 & 59.0 & 67.6 & 37.3 & 73.7&55.3 &64.9& 79.2& 87.5\\
\textbf{+ PGT } &\textbf{96.1} &  \textbf{84.0} & 74.4 & 79.3 & \textbf{67.5} & \textbf{62.9} & 68.4 & 45.3 & 74.2&56.5 &65.4 & 78.1 & 87.5 \\
+ Specialized Mix & 93.7 &82.8 & 70.8 & \textbf{79.9} & 63.4 &  62.7 & 71.5 & 42.0 & \textbf{74.4}&\textbf{56.9}& 65.2& \textbf{79.6} & \textbf{87.6}\\
ViGoRL-3B & 96.2&74.1 & \textbf{78.0} & 77.3& 64.8 &  47.3 & 65.0 & 44.7 & -&37.9 &50.7& 71.2& 86.1 \\
Spatial-Ladder-3B & 88.8 &60.4 & 72.4$^*$ & 74.9$^*$ & 43.1&  52.5 & 58.6& \textbf{48.1} & 68.2&48.9& 52.0 & 73.5 & 84.7\\
\midrule
Qwen2.5-VL-7B & 96.8 &83.8 & 77.7 & 83.5 & 72.4 &  67.5 & 72.8 & 53.3 & 76.4&62.2 & 65.8& 82.7 & 87.4\\
\textbf{+ PGT} & 98.0 &\textbf{85.7} & 78.2 & \textbf{84.6} & 73.5  &\textbf{69.3} & 74.8& 54.0 &76.4&63.5& 67.1& 83.2 & 88.2 \\
+ Specialized Mix & 96.4&85.5& \textbf{78.7} & 82.5 & 72.4&  68.9& 74.7 & 52.0& \textbf{76.6}&62.5& 66.2& 82.7 & 86.8\\
Image Jigsaw& 97.4& 85.4& 77.8 & 83.0 & 70.4& 68.5& 73.7 & \textbf{58.0}& 75.9&62.9& 66.2& 82.9 & 87.5\\
ThinkLite-VL& \textbf{98.3}& 83.3& 76.6 & 80.5 & \textbf{73.9}& 67.5& 75.9& 32.0& 64.3&72.3& \textbf{71.0}& \textbf{83.4} & \textbf{88.5}\\

\bottomrule
\end{tabular}}
\end{sc}
\end{small}
\end{center}
\vskip -0.1in
\end{table*}\looseness-1

In this subsection, we aim to showcase that PGT finetuning on state-of-the-art MLLMs is an effective way to improve their fine-grained visual understanding. To do so, we use PGT to finetune a suite of state-of-the-art MLLMs, including Qwen-2.5-VL (3B, 7B)~\citep{bai2025qwen25vltechnicalreport}, LLaVA-Next (Vicuna-1.5-7B, Llama3-8B)~\citep{li2025spatialladderprogressivetrainingspatial}, and InternVL3-8B~\citep{zhu2025internvl3exploringadvancedtraining}. In this case, to avoid assumptions about real data access, we built PGT on uniformly gray images.\looseness-1
\paragraph{Experimental Setup.}
We conduct controlled  supervised fine-tuning experiments where each model is trained for 2 epochs with a learning rate of $1 \times 10^{-4}$. For these experiments, we utilize LoRA-based~\citep{hu2021loralowrankadaptationlarge} fine-tuning on a set of 5k  PGT samples rendered on neutral gray backgrounds; we do not use any real data and assume that all images $I$ are uniformly gray and $\mathcal{S} = \emptyset$. Similar to our previous experimental settting~\ref{sec:mm_train}, we uniformly sample a PGT task for each of the 5k samples. We use the same evaluation benchmarks as in subsection~\ref{sec:mm_train}. Exact hyperparameters for this section are given in Appendix~\ref{app:mllm_ft_hyper}.\looseness-1

\paragraph{Results.}
Results are presented in Table~\ref{tab:sota_results}. As shown in the table, the application of PGT to advanced models like Qwen-2.5-VL and InternVL3-8B consistently yields performance gains, particularly in vision-centric benchmarks that demand high-precision localization and counting. In particular, we observe that the LLaVA-Next-7B model experiences a substantial boost of +8.3\% on CV-Bench-2D , +6.2\% on VSR and +14.9\% on TallyQA after PGT fine-tuning. Importantly, finetuning on PGT does not hurt the performance of models trained on data mixes targeting improved fine-grained understanding like InternVL3-8B, and even improves the performance of the model on most vision-centric benchmarks -- \eg with boosts +1.1\% on CV-Bench-2D and +3.3\% on RealWorldQA. These results indicate that the spatial primitives learned from geometric overlays are not redundant to the large-scale pre-training of state-of-the-art models but can rather provide a complementary specialized grounding signal that enhances their fine-grained perception.

A key question is whether PGT data is substantially less effective than specialized real-world data for teaching fine-grained understanding. A significant challenge in vision-centric training is the inherent noise in human-annotated datasets. To quantify the effectiveness of our PGT compared to annotated specialized datasets, we construct a \emph{Specialized Mix} that mimics the taxonomy of PGT. To do so, we source samples from the training sets of TallyQA (counting), VSR (relational), and the distance-estimation subset of the Spatial-Ladder-26k dataset~\citep{li2025spatialladderprogressivetrainingspatial}. This mix is created to be the same size as our PGT -- \ie 5k samples. We use the constructed specialized mix to finetune Qwen2.5-VL-3B and Qwen2.5-VL-7B, and compare their performance to that of the same models finetuned on PGT-augmented data. The results of the controlled baseline trained on the above-described Specialized Mix of real data are designated as \texttt{+Specialized mix} in Table~\ref{tab:sota_results}. When comparing PGT results to those of the Specialized Mix, we observe that PGT achieves higher results on the vast majority of relational reasoning, vision-centric and 3D benchmarks. In particular, even if some evaluations are in-distribution \wrt the specialized mix, PGT leads to a +3.6\% (compared to +2.4\% for the specialized mix) boosts in VSR on the Qwen2.5-VL-3B backbone. Interestingly, we observe that gains related to the 2D relative distance estimation PGT are comparable to those achieved by adding real world 3D relative distance estimation data in the Specialized Mix, leading to a +7.8\% boost for PGT compared to a +8.4\% boost using the Specialized Mix on Qwen-2.5-VL-3B. It is worth noting that neither PGT nor the specialized mix result in excessive degradation of general perception benchmarks.\looseness-1

We further compare the Qwen2.5-VL models finetuned with PGT data against other specialized models, which are specifically post-trained to improve finegrained understanding. More specifically, we compare Qwen2.5-VL-3B to its ViGoRL-3B~\citep{sarch2025groundedreinforcementlearningvisual} and Spatial-Ladder-3B~\citep{li2025spatialladderprogressivetrainingspatial} counterparts. ViGoRL consists of a post-training pipeline  designed to encourage the model to incorporate spatial coordinates grounding (x,y) within its language thought. Spatial-Ladder designs a progressive RL training framework using a specialized dataset. Similarly, we compare Qwen2.5-VL-7B to Image-Jigsaw-7B~\citep{wu2025visualjigsawposttrainingimproves} and Thinklite-VL~\citep{wang2025sotalessmctsguidedsample}. Image-Jigsaw consists of an RFT phase leveraging self-supervised jigsaw puzzle tasks, and Thinklite-VL focuses on high quality data selection based on the reasoning iterations to do RL finetuning. Notably, despite our significantly smaller fine-tuning budget (5k samples vs. larger specialized corpora) and no human annotations, we observe that our PGT finetuning can be as effective or even more effective than a specialized training dataset on relational reasoning, and vision-centric and 3D tasks.\looseness-1

\section{Ablations}\label{sec:ablations}
In this section, we aim to systematically characterize which components of the proposed PGT training lead to the observed improvements. To do so, we run the ablations on a single backbone, namely Llama-3-8B~\cite{grattafiori2024llama3herdmodels}, and follow the experimental setup of subsection~\ref{sec:mm_train} leveraging the LLava-Instruct-1.5 dataset~\citep{liu2023visualinstructiontuning}.
\looseness-1 
To rigorously assess how procedurally generated tasks transfer to specific vision-centric domains, we group our benchmarks into four primary axes of evaluation:\looseness-1
\begin{itemize}[leftmargin=*] 
\item \textbf{Relational Reasoning:} Evaluates spatial binding and relative positioning. We report average results across \textit{What’s Up-COCO}~\cite{kamath2023whatsupvisionlanguagemodels}, \textit{VSR}~\cite{liu2023visualspatialreasoning}, the relational subset of \textit{CV-Bench-2D}~\cite{tong2024cambrian1fullyopenvisioncentric}, and the spatial splits of \textit{SEED}~\cite{li2023seedbenchbenchmarkingmultimodalllms} and \textit{BLINK}~\cite{fu2024blinkmultimodallargelanguage}.\looseness-1 \item \textbf{Counting:} Focuses on precise object enumeration. We average results across \textit{TallyQA}~\cite{acharya2018tallyqaansweringcomplexcounting} and the counting splits of \textit{MMSTAR}~\cite{chen2024rightwayevaluatinglarge}, \textit{SEED}, \textit{CV-Bench}, and \textit{BLINK}.\looseness-1 \item \textbf{3D/Depth Understanding:} Probes implicit depth perception and 3D geometry using \textit{CV-Bench-3D} and \textit{BLINK (depth)}. We report average results across these benchmarks.\looseness-1 \item \textbf{General Perception:} Ensures no degradation in standard VQA or hallucination metrics. We average results from \textit{GQA}~\cite{DBLP:journals/corr/abs-1902-09506}, \textit{AI2D}~\cite{Hiippala_2020}, \textit{POPE}~\cite{li2023evaluatingobjecthallucinationlarge}, \textit{RealWorldQA}~\cite{xai2024realworldqa}, \textit{MMVP}, and general aggregated results of \textit{SEED} and \textit{MMSTAR}.\looseness-1
\end{itemize}

We perform a set of ablations that isolate critical design choices that we made for our PGT training: the impact of each task in PGT (subsection~\ref{sec_abla_task}), the impact of PGT on increasing real data sizes(subsection~\ref{sec_abla_scale}), and finally the impact of the proportion of PGT-augmented data \wrt to the base instruction tuning real training data(subsection~\ref{sec_abla_prop}). Exact hyperparameters for this section are detailed in Appendix~\ref{app:mllm_ft_hyper}.\looseness-1

\subsection{The impact of each task in PGT}\label{sec_abla_task}
To evaluate the specific contribution of each task in PGT, we conduct a leave-one-out ablation study across tasks, and compare results with the baseline leveraging regular finetuning (reg-FT) and with full PGT instruction tuning. We compute results by averaging performance across the benchmarks targeting different skills and report them in Table~\ref{tab:ablation_tasks}. We note that the removal of targeted tasks significantly degrades the corresponding skill axis. For example, removing the relational understanding PGT results in a 8.3\% drop on average in relational understanding, while removing the counting PGT results in a 1.2\% performance decrease in counting tasks. Importantly, removing any PGT does not result in notable variations in general understanding tasks. We also observe that excluding the relative distance estimation task causes the 3D/Depth understanding score to substantially reduce from 70.9\% to 61.7\%, nearly matching the 60.8\% baseline performance. This empirical result demonstrates that the model’s ability to compare relative 2D distances between abstract primitives serves as the foundational primitive for emergent relative object depth comparison and 3D understanding. We also observe that the exclusion of the analogy task does not consistently degrade performance across different axes, suggesting that direct spatial and quantitative grounding provide the most robust transfer signals for the evaluated benchmarks.\looseness-1
\begin{table}[h]
\centering
\caption{Leave-one-out task ablation: performance (\%) when removing specific procedurally generated tasks from the full PGT suite. All MLLM are based on Llama-3-8B LLM backbone.\looseness-1}
\label{tab:ablation_tasks}
\begin{small}
\resizebox{\columnwidth}{!}{\begin{tabular}{lcccc}
\toprule
\textbf{Configuration} & \textbf{Relational} & \textbf{Counting} & \textbf{3D/Depth} & \textbf{General} \\
\midrule
Baseline (reg-FT) & 65.2 & 52.5 & 60.8 & 61.9 \\
\textbf{Full PGT} & \textbf{75.7} & \textbf{55.2} & \textbf{72.9} & \textbf{62.2} \\
\midrule
w/o Relational & 66.0 & 54.0 & 68.4 & 61.7 \\
w/o Counting & 73.9 & 51.5 & 69.3 & 61.9 \\
w/o Distance & 72.8 & 53.8 & 61.7 & 61.9 \\
w/o Analogy & 76.0 & 52.0 & 69.2 & 61.5 \\
\bottomrule
\end{tabular}\looseness-1}
\end{small}
\end{table}

\subsection{The benefits of PGT when scaling real data sizes}\label{sec_abla_scale}
We validate whether the benefits of PGT persist as the size of the underlying real dataset increases or whether PGT is mostly beneficial in the low instruction tuning data regime. To do so, we propose to compare a MLLM model based on Llama-3-8b trained at the size of the Llava-Instruct-1.5 dataset (with $\sim$0.6M samples) to the same model trained on real data from Cambrian-7M~\citep{tong2024cambrian1fullyopenvisioncentric} training set (with $\sim$7M). Both models are compared by considering instruction tuning with and without PGT data. As shown in Table~\ref{tab:ablation_scale}, scaling the real dataset from 0.6M samples to 7M samples results in performance improvements across different tasks. Leveraging PGT provides further boosts, notably increasing the performance of the MLMM trained on Cambrian-7M by +6.1\% to +7.8\% on relational and depth reasoning tasks, respectively. This finding confirms that the current bottleneck for fine-grained reasoning in MLLMs is not the sheer volume of semantic instruction tuning data, but rather the density of high-quality spatial supervision. Our ablation shows that PGT effectively addresses this bottleneck regardless of the real (semantic) data training scale.\looseness-1
\begin{table}[h]
\centering
\caption{Impact of PGT across different real, semantic training dataset scales (0.6M vs. 7M dataset sizes). MLLM is based on Llama-3-8b LLM backbone.\looseness-1}
\label{tab:ablation_scale}
\resizebox{\columnwidth}{!}{%
\begin{tabular}{lcccc}
\toprule
\textbf{Dataset Scale} & \textbf{Relational} & \textbf{Counting} & \textbf{3D/Depth} & \textbf{General} \\
\midrule
0.6M (Baseline, reg-FT) & 65.2 & 52.5 & 60.8 & 61.9 \\
\textbf{0.6M + PGT} & \textbf{75.7} & \textbf{55.2} & \textbf{72.9} & \textbf{62.2} \\
\midrule
7M (Baseline, reg-FT) & 69.7 & 57.5 & 62.5 & 64.9 \\
\textbf{7M + PGT} & \textbf{75.8} & \textbf{59.4} & \textbf{70.3} & \textbf{65.8} \\
\bottomrule
\end{tabular}%
}
\end{table}
\begin{table}[h!]
\centering
\caption{Impact of the proportion of PGT-augmented samples on MLLM's performance. MLLM is based on Llama-3-8b LLM backbone. Real data from lava-Instruct-1.5 used for captioning alignment and instruction tuning. When highlighted, PGT-augmented data is used for instruction tuning.\looseness-1}
\label{tab:ablation_proportion}
\resizebox{\columnwidth}{!}{%
\begin{tabular}{lcccc}
\toprule
\textbf{Training} & \textbf{Relational} & \textbf{Counting} & \textbf{3D/Depth} & \textbf{General} \\
\toprule
Baseline (reg-FT) & 65.2 & 52.5 & 60.8 & 61.9 \\ \midrule
PGT 100\% & \textbf{75.7} & 55.2 & \textbf{72.9} & \textbf{62.2} \\
PGT 50\% & 74.3 & \textbf{57.1} & 71.2 & 62.1 \\
PGT 10\% & 74.2 & 56.4 & 70.3 & 61.7 \\
PGT 5\% & 73.8 & 53.0 & 69.6 & 61.9 \\
\bottomrule
\end{tabular}%
}
\end{table}\looseness-1
\subsection{The impact of the proportion of PGT-augmented data \wrt real data}\label{sec_abla_prop}
We ablate the impact of the overall proportion of PGT-augmented data on the MLLM's performance. To do so, we consider applying PGT on 100\%, 50\%, 10\% and only 5\% of the real, semantic training data from Llava-Instruct-1.5, and train MLLMs based on Llama-3-8b LLM backbone. Results are presensed in Table~\ref{tab:ablation_proportion}. We observe that applying PGT, even to small amounts of data, results in substantial performance improvements across different skills, with boosts being more pronounced in relational and 3D/Depth understanding. Although the benefits are stronger when applying PGT to at least 50\% of the data, most of the gains are already realized when leveraging PGT on the smallest fraction of data considered (5\%) -- with improvements of over 8\% in relational and 3D/Depth understandings. Counting is the only skill that appears to notably benefit from additional PGT augmentations during training. Importantly, the performance on general benchmarks remains roughly constant across different proportions of PGT-augmented data used for instruction tuning, and is comparable to the performance of the baseline trained on non-augmented data.\looseness-1

\section{Conclusion and Future Work}
\paragraph{Conclusion.} In this work, we introduced Procedurally Generated Tasks (PGT) designed to tackle finegrained compositional understanding of MLLMs. We leverage PGT to train or finetune MLLMs, pushing the model to not rely on linguistic priors and semantic shortcuts to answer visual question. By forcing models to reason over unambiguous geometric primitives, we encourage the visual attention mechanism to ground itself in the actual image content rather than falling back on statistical correlations in the text. Through extensive experimentation, including both instruction tuning of MLLMs and finetuning of state-of-the-art MLLMs, we showed that leveraging PGT-augmented data not only transfers to real-world images but consistently boosts relational, quantitative, and 3D/Depth understanding, while maintaining general perception capabilities of the models. Moreover, we observed that 3D capabilities may emerge from unambiguous geometric primitives overlaid on real 2D data, highlighting the effectiveness of procedurally generated data and suggesting that this data holds the potential to address the existing bottleneck of fine-grained perception without requiring architectural modification or massive data scaling.\looseness-1

Beyond immediate performance gains, PGT serve a critical role as a low-cost diagnostic instrument. The design space for MLLMs is combinatorially complex—spanning architecture choices, training stages, and data curation—, making it difficult to pinpoint whether a failure stems from a lack of visual capacity or a lack of alignment. Echoing the findings of~\citet{fu2025hiddenplainsightvlms}, who showed that visual representations often contain details that the language model simply ignores, our work confirms that the bottleneck for fine-grained reasoning is often not the visual encoder's resolution or the LLM's size, but the training signal itself. PGT provides a low-cost and effective method to validate the influence of this bottleneck: if a model's performance spikes with PGT, the capability was likely present but dormant, waiting for the right grounding signal.\looseness-1

\textbf{Limitations and Future Work.  }  We position PGT not as a replacement for data scaling or architectural enhancements, but as an orthogonal and highly efficient complementary signal. While model-centric approaches focus on extending a model’s theoretical capacity (\eg via higher resolution or larger backbones), PGT addresses the alignment bottleneck that prevents models from accessing that capacity. Given the substantial performance gains observed across diverse benchmarks with negligible computational overhead, we argue that PGT is justified as a standard integration in modern training recipes, regardless of the underlying architecture.

Consequently, PGT should also serve as a necessary diagnostic baseline. Before attributing spatial failures to "insufficient resolution" or "weak vision backbones," we encourage researchers to first verify if the model can solve PGT. Success on PGT indicates that the necessary visual features are present but dormant, waiting for a clearer training signal to override linguistic priors. We acknowledge, however, that the specific suite of tasks introduced in this paper is a proof-of-concept rather than an exhaustive catalog. While our current suite is not meant to be a comprehensive list of all possible tasks, we provide a detailed discussion of promising extension avenues for the PGT framework---including 3D primitives, temporal understanding, and multi-step reasoning---in Appendix~\ref{app:extended_future_work}.
\looseness-1

\section*{Impact Statement}
This paper presents work whose goal is to advance the field of machine learning. In particular, this paper focuses on improving fine-grained spatial understanding. There are many potential societal consequences of our work, none of which we feel must be specifically highlighted here.

\nocite{langley00}

\bibliography{example_paper}
\bibliographystyle{icml2026}


\newpage
\appendix
\onecolumn

\section{Additional Dataset details}\label{app:dataset}

\begin{figure*} [!h]
    \centering
    \includegraphics[width=\textwidth]{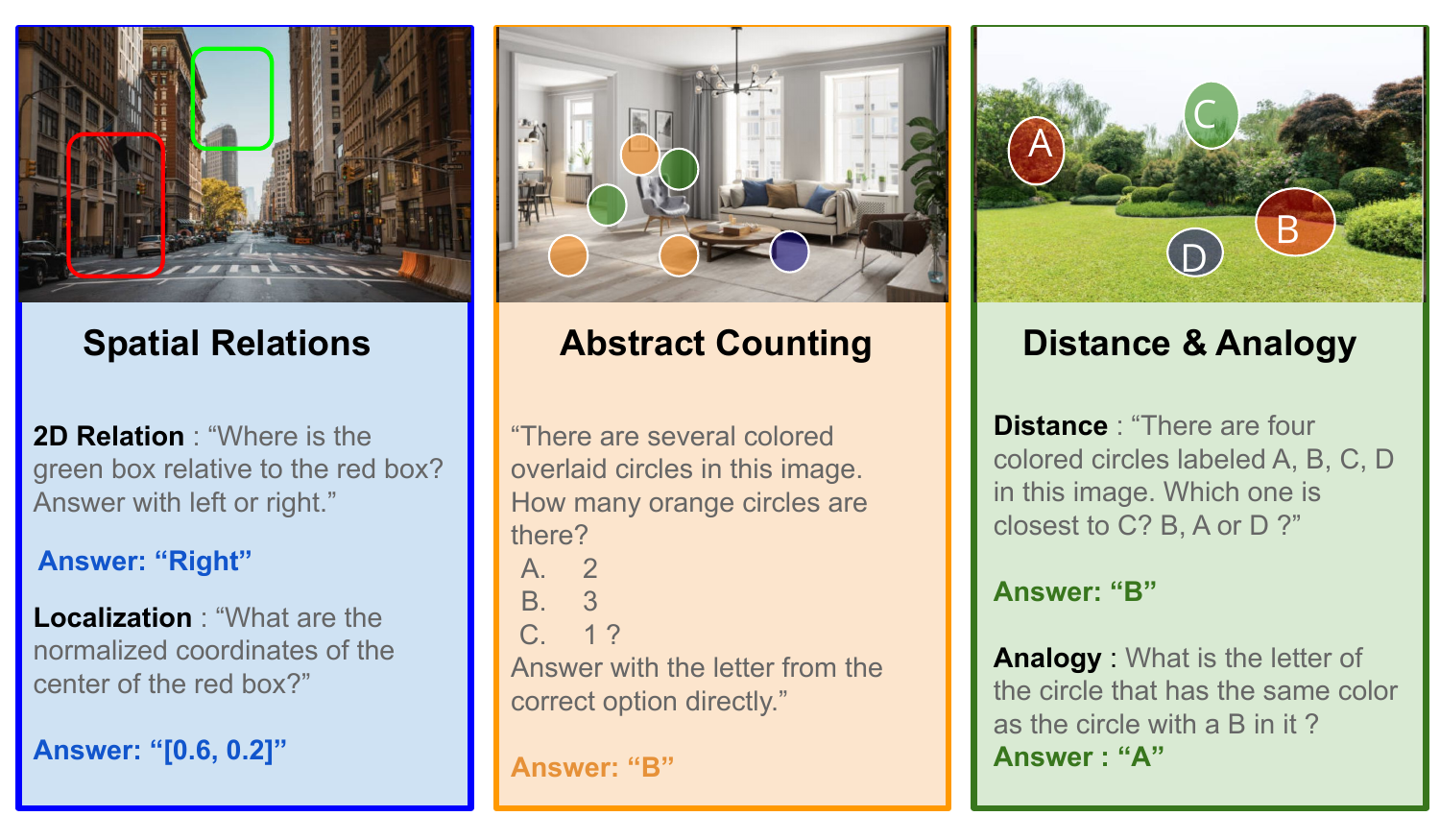}
   \caption{Our suite of PGT: (left) spatial relationship reasoning, (center) abstract counting, and (right) 2D relative distance estimation.\looseness-1}
    \label{fig:pgt_tasks}
\end{figure*}

In this section, we provide the specific prompts and templates used for generating the Procedurally Generated Tasks (PGT) as well as the prompts used for the Specialized Mix (constructed from TallyQA, VSR, and Spatial Ladder).

\subsection{Handling Occlusion and Semantic Preservation}
\label{app:occlusion}
To mitigate the potential occlusion of critical semantic features in the original image, specific geometric primitives---such as the colored circles used in the counting tasks---are explicitly rendered as semi-transparent during our procedural generation pipeline. Our empirical results validate that this theoretical interference does not negatively manifest in practice. As demonstrated in Section~\ref{sec:mm_train}, performance on general perception benchmarks (e.g., MMSTAR, GQA, and POPE) remains stable or fluctuates minimally across all model backbones when PGT is applied. This confirms that the semantic integrity of the original image is sufficiently preserved, allowing the model to effectively learn from the original instruction-tuning QA pairs alongside the PGT augmentations.

\subsection{Procedurally Generated Tasks (PGT) Prompts}

For each task in our PGT suite, we utilize a standardized prompt template to ensure consistency. The geometric primitives (e.g., bounding boxes, colored circles) are overlaid on the image prior to feeding it into the model. We varied templates with and without formatting instructions. When several template are available one is sampled uniformly at random.

\paragraph{Spatial Relationship Understanding.} This task involves two overlaid bounding boxes (typically green and red). The model is asked to identify the spatial relationship of the green box relative to the red box. The model is also tasked to predict the center coordinates of one the boxes as shown in Figure~\ref{fig:pgt_tasks}.
\begin{tcolorbox}[colback=blue!5!white,colframe=blue!75!black,title=Spatial Relationship Prompt]
\textbf{Template A:} \texttt{USER}: Where is the [color\_A] box relative to the [rel\_B] box? Answer with [rel\_A] or [rel\_B]. \texttt{GPT} : [rel\_A]\\

\textbf{Template B:} \texttt{USER}: Based on the image, is this statement True or False? The [color\_A] box is [rel\_A] of [color\_B] box? Answer with True or False directly. \texttt{GPT} : True \\

\textbf{Template C:} \texttt{USER} :Considering the relative positions of the [color\_A] box and the [color\_B] box in the image provided, where is the [color\_A] box  located with respect to the [color\_A] box? Select from the following choices. (A) [rel\_A]  (B) [rel\_B] \texttt{GPT} : [rel\_A] 
\end{tcolorbox}

\paragraph{Counting.} In this task, semi-transparent circles of various colors are projected onto the scene. The model must count only the circles of a specific target color.
\begin{tcolorbox}[colback=orange!5!white,colframe=orange!75!black,title=Counting Prompt]

\textbf{Template A :} \texttt{USER} : How many [color] overlay circles are there? Answer with the number directly \texttt{GPT} : [num] \\ 

\textbf{Template B :} \texttt{USER} : How many [color] overlay circles are there? (A) [option\_A] (B) [option\_B] (C) [option\_C] Answer with the option's letter from the given choices directly \texttt{GPT} : [letter]
\end{tcolorbox}

\paragraph{Analogy and Relative Distance.} This task utilizes multi-point overlays labeled with identifiers (e.g., A, B, C, D). The model is prompted to estimate distances (e.g., finding the closest point to a target) or solve a simple color-based analogy.
\begin{tcolorbox}[colback=green!5!white,colframe=green!75!black,title=Relative Distance Prompt]
\textbf{Template Distance:} \texttt{USER}: There are four colored circles labeled A, B, C, D in this image. Which one is closes to [target\_letter] : [option\_A], [option\_B], or [option\_C] ? \texttt{GPT} : [letter] \\

\textbf{Template Analogy: } \texttt{USER}: What is the letter of the circle that has the same color as the circle with a [target\_letter] in it ? \texttt{GPT}: [letter]

\end{tcolorbox}

\subsection{Specialized Training Mix Prompts}

To compare PGT against real-world annotated data, we constructed a Specialized Mix using samples from TallyQA~\citep{acharya2018tallyqaansweringcomplexcounting}, VSR~\citep{liu2023visualspatialreasoning}, and Spatial Ladder-26k~\citep{li2025spatialladderprogressivetrainingspatial}. When feasible, we standardized the answer formatting for these datasets to align with our PGT  format and diversity.

\paragraph{Counting (TallyQA).} Counting taks are sourced from TallyQA~\citep{acharya2018tallyqaansweringcomplexcounting}, like our PGT set we alternate between free form answers and multiple choice. We only vary the answering format based on the answer. The question is taken from the original dataset directly.
\begin{tcolorbox}[colback=orange!5!white,colframe=orange!75!black,title=TallyQA Prompt (Specialized Mix)]

\textbf{Template A :} \texttt{USER} : How many [obj] are there? Answer with the number directly \texttt{GPT} : [num] \\ 

\textbf{Template B :} \texttt{USER} : How many [obj] are there? (A) [option\_A] (B) [option\_B] (C) [option\_C] Answer with the option's letter from the given choices directly \texttt{GPT} : [letter]
\end{tcolorbox}

\paragraph{Relational Reasoning (VSR).} For relational tasks sourced from Visual Spatial Reasoning (VSR)~\citep{liu2023visualspatialreasoning}, we use true/false validation prompts regarding the spatial configuration of objects. Note that VSR employs more relationships than our PGT set which is restricted 2D to left/right, above/below spatial statements.
\begin{tcolorbox}[colback=gray!5!white,colframe=gray!75!black,title=VSR Prompt (Specialized Mix)]
\textbf{Template:} \texttt{USER}: Based on the image, is this statement True or False? [statement] Answer with True or False directly. \texttt{GPT} : True
\end{tcolorbox}

\paragraph{Distance Estimation (Spatial Ladder).} For distance estimation, we utilize the distance-estimation subset from the Spatial-Ladder-26k 
dataset~\citep{li2025spatialladderprogressivetrainingspatial} without changing the original formatting of the answer.
\begin{tcolorbox}[colback=gray!5!white,colframe=gray!75!black,title=Spatial Ladder Prompt (Specialized Mix)]

\textbf{Example:}  \texttt{user} : Measuring from the closest point of each object, which of these two objects (tv, pillow) is closer to the table? A. tv B. pillow Answer with the option's letter from the given choices directly. \texttt{gpt} : A.
\end{tcolorbox}

\section{Additional Training Details}\label{app:training}
\label{app:training_details}

In this section, we provide the detailed hyperparameters and training configurations used for our experiments. We categorize our experiments into two main settings: (1) Instruction Tuning and Ablations (Sections~\ref{sec:mm_train} and \ref{sec:ablations}), which utilize the Prismatic~\citep{karamcheti2024prismaticvlmsinvestigatingdesign} training framework, and (2) Finetuning of State-of-the-Art (SOTA) MLLMs (Section~\ref{sec :finetuning}), which utilizes the LLaMA-Factory framework.

\subsection{Instruction Tuning and Ablations }\label{app:mllm_training_hyper}

For the multimodal training experiments (Section 4.1) and the ablation studies (Section 5), we adopted the Prismatic training framework~\citep{karamcheti2024prismaticvlmsinvestigatingdesign}. We followed a two-stage training protocol: first, pre-training the adapter for captioning alignment, followed by instruction tuning of both the adapter and the LLM backbone while keeping the vision encoder frozen. We adopt a 2 layer MLP for the adapter and fix the vision backbone to the most commonly used CLIP-L-14-336px.

Following \citet{karamcheti2024prismaticvlmsinvestigatingdesign} the captioning alignment training is done with a global batch size of 256 and a learning rate of $1e-3$. The instruction tuning training is done with a global batch size of 128 and a learning rate of $2 \times 10^{-5}$,  a cosine schedule and a 0.1 warmup . Tables~\ref{tab:llava_opt} and \ref{tab:llava_config} detail the specific optimization and system hyperparameters used in this setting.

The ablation experiment measuring the impact of the multimodal training data~\ref{sec_abla_prop} using the Cambrian-7M dataset follows the hyperparameters given in \citet{tong2024cambrian1fullyopenvisioncentric}, we also detail them in Tables~\ref{tab:cambrian_opt} and \ref{tab:cambrian_config}.\looseness-1
\begin{table}[h]
    \centering
    \footnotesize
    \renewcommand{\arraystretch}{0.9}
    \setlength{\tabcolsep}{4pt}
    
    \begin{minipage}[t]{0.48\textwidth}
        \centering
        \caption{Optimization Hyperparameters (LLaVA-Instruct-1.5).}
        \label{tab:llava_opt}
        \vspace{0.1cm}
        \begin{tabular}{lcc}
            \toprule
            \textbf{Hyperparameter} & \textbf{Stage 1} & \textbf{Stage 2} \\
            \midrule
            Optimizer & AdamW & AdamW \\
            Learning Rate & 1e-3 & 2e-5 \\
            LR Schedule & Cosine & Cosine \\
            Warmup Ratio & 0.03 & 0.1 \\
            Global Batch Size & 256 & 128 \\
            Weight Decay & 0.0 & 0.1 \\
            Grad. Clipping & 1 & 1 \\
            \bottomrule
        \end{tabular}
    \end{minipage}
    \hfill
    \begin{minipage}[t]{0.48\textwidth}
        \centering
        \caption{Training \& System Config (LLaVA-Instruct-1.5).}
        \label{tab:llava_config}
        \vspace{0.1cm}
        \begin{tabular}{lcc}
            \toprule
            \textbf{Config} & \textbf{Stage 1} & \textbf{Stage 2} \\
            \midrule
            Epochs & 1 & 2 \\
            Precision & BF16 & BF16 \\
            Max Seq. Len & 2048 & 2048  \\
            Num GPUs & 8 & 16 \\
            Sharding & hybrid\_shard\_zero2 &hybrid\_shard \\
            \bottomrule
        \end{tabular}
    \end{minipage}
\end{table}
\begin{table}[h]
    \centering
    \footnotesize
    \renewcommand{\arraystretch}{0.9}
    \setlength{\tabcolsep}{4pt}
    
    \begin{minipage}[t]{0.48\textwidth}
        \centering
        \caption{Optimization Hyperparameters (Cambrian).}
        \label{tab:cambrian_opt}
        \vspace{0.1cm}
        \begin{tabular}{lcc}
            \toprule
            \textbf{Hyperparameter} & \textbf{Stage 1} & \textbf{Stage 2} \\
            \midrule
            Optimizer & AdamW & AdamW \\
            Learning Rate & 1e-3 & 4e-5 \\
            LR Schedule & Cosine & Cosine \\
            Warmup Ratio & 0.03 & 0.1 \\
            Global Batch Size & 512 & 512 \\
            Weight Decay &0  & 0.1\\
            Grad. Clipping & 1 & 1\\
            \bottomrule
        \end{tabular}
    \end{minipage}
    \hfill
    \begin{minipage}[t]{0.48\textwidth}
        \centering
        \caption{Training \& System Config (Cambrian).}
        \label{tab:cambrian_config}
        \vspace{0.1cm}
        \begin{tabular}{lcc}
            \toprule
            \textbf{Config} & \textbf{Stage 1} & \textbf{Stage 2} \\
            \midrule
            Epochs & 1 & 1 \\
            Precision & BF16 & BF16 \\
            Num GPUs & 8 & 16 \\
            Max Seq. Len & 2048 & 2048 \\
             Sharding & hybrid\_shard\_zero2 &hybrid\_shard \\
            \bottomrule
        \end{tabular}
    \end{minipage}
\end{table}\looseness-1

\subsection{Finetuning SOTA MLLMs }\label{app:mllm_ft_hyper}

For the experiments involving the finetuning of state-of-the-art MLLMs (Section \ref{sec :finetuning}), such as Qwen-2.5-VL, LLava-NexT and InternVL3, we utilized the LLaMA-Factory framework. These models were finetuned using Low-Rank Adaptation (LoRA) on the PGT-augmented data (or the specialized mix). The exact instruction tuning prompt for this data are given in Section~\ref{app:training}.

We trained the models on 5000 samples, for 2 epochs with a learning rate of $1 \times 10^{-4}$. We used a LoRA rank of 8 and alpha of 32 targeting all the keys, queries, and values linear layer of the language model. The multimodal projector and vision backbone are kept frozen. Table~\ref{tab:llama_factory_params} lists the full configuration and hyperparameters.\looseness-1

\begin{table}[h]
    \centering
    \footnotesize
    \renewcommand{\arraystretch}{0.9}
    \setlength{\tabcolsep}{8pt}
    \caption{Hyperparameters for SOTA MLLM Finetuning (LLaMA-Factory).}\looseness-1
    \label{tab:llama_factory_params}
    \vspace{0.1cm}
    \begin{tabular}{lc}
        \toprule
        \textbf{Hyperparameter} & \textbf{Value} \\
        \midrule
        \multicolumn{2}{c}{\textit{Optimization}} \\
        Learning Rate & $1 \times 10^{-4}$ \\
        LR Schedule & Cosine \\
        LR ratio & 0.1 \\
        Optimizer & AdamW \\
        Global Batch Size & 64 \\
        Max Gradient Norm & 2 \\
        \midrule
        \multicolumn{2}{c}{\textit{LoRA Configuration}} \\
        LoRA Rank ($r$) & 8 \\
        LoRA Alpha ($\alpha$) & 32 \\
        \midrule
        \multicolumn{2}{c}{\textit{Training Duration}} \\
        Num Epochs & 2 \\
        Data Sample Size & 5k (PGT/Specialized Mix) \\
        \bottomrule
    \end{tabular}\looseness-1
    \looseness-1
\end{table}\looseness-1

\section{Additional Qualitative Analyses}
\label{app:response_analysis}

\subsection{Identified Spatial Shortcuts}
In this section, we provide a qualitative analysis of the failure modes and reasoning shortcuts observed in current state-of-the-art MLLMs. Echoing the findings of~\citet{fu2025hiddenplainsightvlms}, who showed that visual representations often contain details that the language model simply ignores or overlooks, our work confirms that the bottleneck for fine-grained reasoning is often not the visual encoder's resolution or the LLM's size, but the training signal itself. By examining the models' responses and reasoning traces (for thinking models), we identify recurring non-robust heuristics that models employ to solve spatial tasks.

We specifically observed these common shortcut mechanisms:
\begin{itemize}
    \item \textbf{Scale-Depth Correlation:} The model assumes larger objects are always closer, probably based on real world statistics.
    \item \textbf{2D Verticality vs. Depth:} The model conflates the 2D $y$-coordinate with depth, assuming objects higher in the image plane are further away, regardless of the actual scene geometry.
    \item \textbf{Detailedness vs. Depth} : The model relies on the common fact that more detailed ( and less blurry objects) are usually located closer to the camera.
 
\end{itemize}

Below, we present examples illustrating these behaviors.

\newtcolorbox{analysisbox}[2][]{
    colback=gray!5!white,
    colframe=gray!75!black,
    title=\textbf{#2},
    fonttitle=\bfseries,
    #1
}

\begin{analysisbox}{Scale-Depth Correlation}
    \begin{minipage}{0.35\textwidth}
        \centering
       
        \includegraphics[width=\linewidth]{figures/examples/shortcut_1.pdf} 
        
    \end{minipage}
    \hfill
    \begin{minipage}{0.60\textwidth}
        \small
        \textbf{Model:} VigoRL-3B\\
        \textbf{Question:}Which object is closer to the camera taking this photo, the door (highlighted by a red box) or the books (highlighted by a blue box)?\\
(A) door\\
(B) books\\

Answer only with the single capital letter corresponding to the correct choice. \\

        \vspace{0.2cm}
        \textbf{Model's Thinking Trace):} \\
        \textit{
To determine which object is closer to the camera, let's consider their relative positions. The door is located at (670, 150), and the books are inside the blue box at (350, 350). Since the books are inside the blue box and the blue box is closer to the camera than the door, the books are closer to the camera. Let's verify this by comparing the distances. The door is at (670, 150), and the books are at (350, 350). The distance from the camera to the door is greater than the distance from the camera to the books. Now, let's consider the perspective of the image. \textcolor{blue}{The door appears smaller than the books because it is farther away.} This confirms that the books are indeed closer to the camera. Another factor to consider is the depth of the scene. The books are positioned inside the blue box, which is closer to the camera than the door. This further supports that the books are closer to the camera. To ensure accuracy, let's also consider the lighting and shadows. The books cast a shadow on the floor, indicating they are closer to the light source, which is consistent with them being closer to the camera. \textcolor{blue}{Lastly, let's examine the scale of objects. The books appear larger in the image compared to the door, suggesting they are closer to the camera. This is another indicator that the books are indeed closer}.
}
        
        \vspace{0.2cm}
        \hrule
        \vspace{0.2cm}
        
        \textbf{Analysis:} \\
        The model relies on a \textit{size heuristic}, equating pixel area with proximity. It fails to account for the intrinsic size difference between the objects, demonstrating a lack of genuine depth perception.
    \end{minipage}\looseness-1
\end{analysisbox}\looseness-1

\vspace{0.5cm}

\begin{analysisbox}{2D Verticality/Details vs. 3D Depth}
    \begin{minipage}{0.35\textwidth}
        \centering
       
        \includegraphics[width=\linewidth]{figures/examples/example3.pdf}
        
    \end{minipage}
    \hfill
    \begin{minipage}{0.60\textwidth}
        \small
        \textbf{Model:}VigoRL-3B \\
        \textbf{Question:} \\
        Which object is closer to the camera taking this photo, the door (highlighted by a red box) or the lamp (highlighted by a blue box)?\\
(A) door\\
(B) lamp\\
Answer only with the single capital letter corresponding to the correct choice.
\\
        
        \vspace{0.2cm}
        \textbf{Model Response (Thinking Trace):} \\
        \textit{
To determine which object is closer to the camera, \textcolor{blue}{let's consider their relative positions. The door (highlighted by a red box) is located near the center-left of the image at (400, 300). The lamp (highlighted by a blue box) is positioned higher up and further back in the room at (550, 50). Since the door is closer to the bottom edge of the image and the lamp is higher up, the door appears closer to the camera}. Let's verify this by comparing their distances from the camera's perspective. The door is located at (400, 300), which is closer to the bottom of the image. The lamp is at (550, 50), which is higher up and further back. Considering the perspective, the door at (400, 300) seems closer to the camera than the lamp at (550, 50). To further confirm, let's consider the depth of the scene. The door at (400, 300) is closer to the foreground compared to the lamp at (550, 50). This suggests that the door is indeed closer to the camera. Another factor to consider is the size of the objects. \textcolor{blue}{The door at (400, 300) appears larger and more detailed, indicating that it is closer to the camera than the smaller and farther lamp at (550, 50)}. To ensure accuracy, let's also examine the lighting. The door at (400, 300) is well-lit, suggesting it is closer to the light source, whereas the lamp at (550, 50) is less illuminated, indicating it is farther away from the light source. This further supports the conclusion that the door is closer to the camera. Lastly, let's consider the shadows. The shadow cast by the door at (400, 300) is more pronounced and closer to the camera, reinforcing that the door is indeed closer to the camera compared to the lamp at (550, 50).}\looseness-1
        
        \vspace{0.2cm}
        \hrule
        \vspace{0.2cm}
        
        \textbf{Analysis:} \\
        The model applies a \textit{projective geometry prior} indiscriminately. While often true for objects on a ground plane, this heuristic fails in indoor scenes with various views , where the 2D $y$-coordinate does not map linearly to depth. It also relies on another correlation where more detailed objects are usually closer to the camera. 
    \end{minipage}\looseness-1
\end{analysisbox}\looseness-1

\begin{analysisbox}{Binding Erros}
    \begin{minipage}{0.35\textwidth}
        \centering
       
        \includegraphics[width=\linewidth]{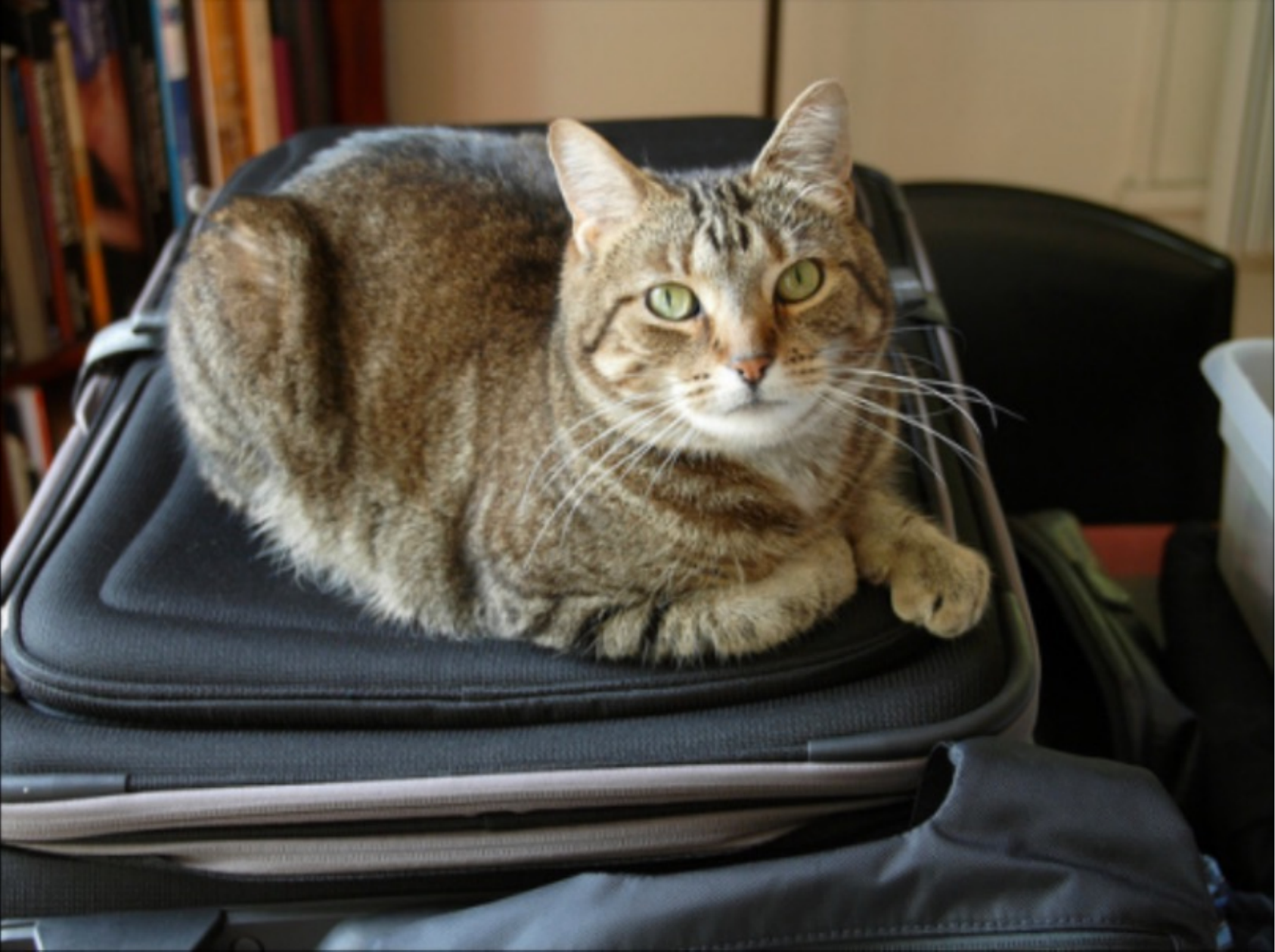}
       
    \end{minipage}
    \hfill
    \begin{minipage}{0.60\textwidth}
        \small
        \textbf{Model:} Spatial-Ladder-3B \\
        
        \textbf{Question:} \\
        Based on the image, is this statement "True" or "False"?The suitcase is under the cat.
Respond with "True" or "False" directly.\\

        \vspace{0.2cm}
        \textbf{Model Response (Thinking Trace):} \\
        \textit{To determine if the statement "The suitcase is under the cat" is true, let's analyze the image. \textcolor{blue}{The cat is positioned on top of the suitcase, not underneath it.} The cat's body is aligned with the top edge of the suitcase, indicating that the cat is on top. (250, 150) To further confirm this observation, let's consider the relative positions of the cat and the suitcase. The cat's head is near the top of the suitcase, and its body extends down towards the bottom of the suitcase. This confirms that the cat is indeed on top of the suitcase, not underneath. (350, 200) Another way to verify this is by examining the shadows and lighting in the image. The shadow cast by the cat appears to fall directly beneath the cat, suggesting that the cat is on top of the suitcase rather than underneath it. (400, 250) To ensure accuracy, let's also look at the texture and material of the objects. The suitcase appears to be made of a hard material, while the cat's fur is soft and fluffy. \textcolor{blue}{This further supports the idea that the cat is on top of the suitcase, not underneath.} (500, 300) Lastly, let's consider the perspective and depth of the image. The cat's position relative to the background elements suggests that it is elevated above the ground, \textcolor{blue}{which aligns with the cat being on top of the suitcase rather than underneath. }(600, 350)} 
        
        \vspace{0.2cm}
        \hrule
        \vspace{0.2cm}
        
        \textbf{Analysis:} \\
        As the reasoning progresses the shifted towards evaluating the statement "The cat is under the suitcase" instead of the initial one "The suitcase is under the cat". It did not remain consistent with the initial relationship binding, switching the subject of the assessed relationship to the most probable statistically (animated subject) in real-world captioned images.
    \end{minipage}\looseness-1
\end{analysisbox}\looseness-1

\renewtcolorbox{analysisbox}[2][]{
    colback=red!5!white,
    colframe=red!75!black,
    title=\textbf{#2},
    fonttitle=\bfseries,
    breakable, 
    #1
}\looseness-1

We also give few example of corrected answer in the CVBench-3D dataset. Answers are taken from LLava-Next-8B where we observed significant boost after finetuning the model on our PGT mix.
Notably, in all those examples, the models systematically predicts that the bigger object is closer to the camera. While this observation is not causal of the observed effect, we want to draw's the reader's attention to the potential shortcuts the model might rely on and how that can inform dataset design.\looseness-1

\begin{analysisbox}{Examples of corrected answers (LLaVA-NeXt-8b)}

    \begin{minipage}{0.30\textwidth}
        \centering
        \includegraphics[width=\linewidth]{figures/examples/example1.pdf} 
      
    \end{minipage}
    \hfill
 \begin{minipage}{0.65\textwidth}
        \small
        \textbf{Question:}   Which object is closer to the camera taking this photo, the books (highlighted by a red box) or the chair (highlighted by a blue box)?\\
(A) books\\
(B) chair\\
Answer only with the single capital letter corresponding to the correct choice.
\\
  
        \textbf{Original Prediction:} B \\
        \textbf{Prediction after PGT ft:} A
    \end{minipage}

    \vspace{0.3cm}
    \hrule
    \vspace{0.3cm}

    \begin{minipage}{0.30\textwidth}
        \centering
        \includegraphics[width=\linewidth]{figures/examples/example2.pdf} 
     
    \end{minipage}
    \hfill
    \begin{minipage}{0.65\textwidth}
        \small
        \textbf{Question:} Which object is closer to the camera taking this photo, the lamp (highlighted by a red box) or the table (highlighted by a blue box)?\\
(A) lamp\\
(B) table\\
Answer only with the single capital letter corresponding to the correct choice.
\\
  
        \textbf{Original Prediction:} B \\
        \textbf{Prediction after PGT ft:} A 
    \end{minipage}

    \vspace{0.3cm}
    \hrule
    \vspace{0.3cm}

    \begin{minipage}{0.30\textwidth}
        \centering
        \includegraphics[width=\linewidth]{figures/examples/example3.pdf} 
   
    \end{minipage}
    \hfill
   \begin{minipage}{0.65\textwidth}
        \small
        \textbf{Question:} Which object is closer to the camera taking this photo, the door (highlighted by a red box) or the lamp (highlighted by a blue box)?\\
(A) door\\
(B) lamp\\
Answer only with the single capital letter corresponding to the correct choice.
\\
  
        \textbf{Original Prediction:} A \\
        \textbf{Prediction after PGT ft:} B 
    \end{minipage}

    \vspace{0.3cm}
    \hrule
    \vspace{0.3cm}

    \begin{minipage}{0.30\textwidth}
        \centering
        \includegraphics[width=\linewidth]{figures/examples/example4.pdf} 

    \end{minipage}
    \hfill
   \begin{minipage}{0.65\textwidth}
        \small
        \textbf{Question:} Which object is closer to the camera taking this photo, the bicycle (highlighted by a red box) or the bus (highlighted by a blue box)?\\
(A) bicycle\\
(B) bus\\
Answer only with the single capital letter corresponding to the correct choice.
\\
  
        \textbf{Original Prediction:} B \\
        \textbf{Prediction after PGT ft:} A
    \end{minipage}

\end{analysisbox}

\section{Additional Ablations}

\begin{table}[h]
\centering
\caption{Impact of the overlay strategy vs. separate synthetic data. All results use the Llama-3-8b backbone.}
\label{tab:ablation_overlay}

\begin{tabular}{lcccc}
\toprule
\textbf{Strategy} & \textbf{Relational} & \textbf{Counting} & \textbf{3D/Depth} & \textbf{General} \\
\midrule
Baseline (reg-FT) & 65.2 & 52.5 & 60.8 & 61.9 \\
\textbf{PGT (Overlay)} & 75.7 & 55.2 & 72.9 & 62.2 \\
Separate Data & 75.8 & 55.3 & 75.4 & 62.8 \\
\bottomrule
\end{tabular}%

\end{table}
In this section we analyze the efficiency of our overlay mechanism compared to a traditional strategy of appending synthetic data as a separate dataset (Separate Data) and thus augmenting the total number of training samples. While training on a separate synthetic mix yields a slight absolute performance advantage—reaching 75.8\% on relational reasoning and 75.4\% on 3D/Depth—it requires a substantial increase in training iterations and computational cost. Our overlay strategy captures the vast majority of these performance gains while maintaining the original training sample count. This demonstrates that overlaying geometric primitives directly onto semantic scenes is not only a computationally superior trade-off but also forces the model to disentangle abstract geometry from real-world features within a single visual context.

\section{Extended Limitations: Task Saturation and Clutter}
\label{app:saturation}

While our main ablation study (Section 5.3) demonstrates that applying PGT to just 5\% of the training data yields significant performance improvements, and scaling up to 100\% maintains stable general perception performance, there is a theoretical upper limit to task density. Specifically, injecting multiple different PGT overlays per image (exceeding a 100\% ratio) introduces the risk of severe image clutter. 

To directly evaluate the impact of exceeding a 100\% ratio, we conducted an additional experiment simulating task accumulation. We finetuned the LLaVA-NeXt-7B model with 1, 2, and 3 distinct PGT tasks overlaid per image sample. The results are summarized in Table~\ref{tab:saturation}.

\begin{table}[h]
\centering
\caption{Impact of task accumulation (injecting multiple PGTs per sample) on LLaVA-NeXt-7B performance.}
\label{tab:saturation}
\begin{tabular}{lcccc}
\toprule
\textbf{Number of PGTs} & \textbf{Relational (\%)} & \textbf{Count (\%)} & \textbf{3D/Depth (\%)} & \textbf{General (\%)} \\
\midrule
\textbf{1 task}  & 69.8 & 53.7 & 61.4 & 63.1 \\
\textbf{2 tasks} & 71.1 & 54.8 & 60.4 & 62.0 \\
\textbf{3 tasks} & 70.8 & 52.5 & 54.8 & 61.1 \\
\bottomrule
\end{tabular}
\end{table}

These results show that while the model can accommodate a moderate increase in task density (peaking at 2 tasks for relational reasoning and counting), pushing the augmentation to 3 tasks causes a noticeable degradation across all metrics, with 3D/Depth understanding experiencing the most severe drop. We acknowledge that this specific setup isolates task accumulation and does not strictly evaluate the occlusion of real-world background information. However, it clearly indicates that injecting too many tasks per image is inherently detrimental to the model's learning capacity, irrespective of background clutter.

\section{Extended Future Work}
\label{app:extended_future_work}

While the tasks presented in the main text serve as a foundational proof-of-concept, the design space for Procedurally Generated Tasks (PGT) is vast. We outline several highly promising avenues for extending the visual PGT framework to address more complex modalities and advanced spatial logic.

\paragraph{3D Geometric Primitives.} 
As demonstrated in Section~\ref{sec:mm_train}, our 2D relative distance tasks elicited emergent 3D depth perception, strongly supporting the hypothesis that these tasks reinforce a shared distance estimation circuit. A natural and highly relevant extension is to explicitly target 3D spatial understanding by procedurally generating 3D geometric primitives. Injecting rendered shapes---such as cubes or spheres with simulated lighting, shadows, and projective perspective---directly onto images holds the potential to push models' 3D understanding significantly further than 2D abstractions alone.

\paragraph{Temporal Reasoning in Video.} 
Extending the PGT framework beyond static images to the video modality represents a critical next step. Procedural geometric tasks focusing on temporal reasoning---such as tracking a procedurally generated marker across frames or predicting the trajectory of a moving overlay---could provide the dense, unambiguous supervision required to improve object permanence and temporal grounding in MLLMs.

\paragraph{Multi-Step Reasoning and Advanced Spatial Logic.} 
Our current PGT suite focuses on fixing the foundational visual understanding layer (\eg binary spatial predicates, basic counting, and depth estimation). However, recent evaluation benchmarks such as ReasonMap~\citep{feng2026reasonmapfinegrainedvisualreasoning}, MapBench~\citep{xing2025largevisionlanguagemodels}, and V*~\citep{wu2023vguidedvisualsearch} demonstrate that MLLMs still heavily struggle with complex visual reasoning and planning, such as multi-hop route finding on topological maps or systematic visual search. We hypothesize that PGT can act as a bridge between basic visual grounding and these complex reasoning capabilities. Future task designs could introduce multi-step geometric pathfinding between labeled nodes or require intersection area estimation of overlapping semi-transparent shapes. Applying the PGT methodology to these complex reasoning tasks via step-by-step prompt expansion (akin to chain-of-thought, but explicitly grounded in procedurally generated visual anchors) is a natural and exciting evolution of this work.


\end{document}